\documentclass[sigconf]{acmart}
\usepackage{multirow}
\usepackage{changepage}
\usepackage{graphicx}
\usepackage{float}
\usepackage{makecell}
\usepackage{booktabs}
\usepackage{tabularx}
\usepackage{enumitem}
\usepackage{amsmath}
\usepackage{mathtools}
\usepackage{amsthm}
\usepackage{threeparttable}
\usepackage[table]{xcolor}
\usepackage{colortbl}
\usepackage{enumitem}
\AtBeginDocument{%
  }

\copyrightyear{2026}
\acmYear{2026}
\setcopyright{cc}
\setcctype{by}
\acmConference[MM '26] {Proceedings of the 34th ACM International Conference on Multimedia}{November 10--14, 2026}{Rio de Janeiro, Brazil}
\acmBooktitle{Proceedings of the 34th ACM International Conference on Multimedia (MM '26), November 10--14, 2026, Rio de Janeiro, Brazil}
\acmISBN{979-8-4007-2213-4/2026/11}
\acmDOI{10.1145/3767308.3835927}
\begin{document}

\title[Towards Reliable Stain Transfer]{Towards Reliable Stain Transfer: An Iterative Data-Model Co-Optimization Framework Based on Multimodal Expert-Guided Assessment}

\author{Siyuan Xu}
\affiliation{
  \institution{East China Normal University}
  \city{Shanghai}
  \country{China}}
\email{syxu@stu.ecnu.edu.cn}

\author{Yan Wang}
\authornote{Corresponding authors.}
\affiliation{
  \institution{East China Normal University}
  \city{Shanghai}
  \country{China}}
\email{ywang@cee.ecnu.edu.cn}

\author{Haofei Song}
\affiliation{
  \institution{East China Normal University}
  \city{Shanghai}
  \country{China}}
\email{hfsong@stu.ecnu.edu.cn}

\author{Lili Gao}
\affiliation{
  \institution{Ruijin Hospital, Shanghai Jiao Tong University School of Medicine}
  \city{Shanghai}
  \country{China}}
\email{gll12216@rjh.com.cn}

\author{Jiansheng Wang}
\affiliation{
  \institution{Hangzhou Hyperspectral Imaging Technology Co., Ltd.}
  \city{Hangzhou}
  \country{China}}
\email{jswang@cee.ecnu.edu.cn}

\author{Qing Zhang}
\affiliation{
  \institution{East China Normal University}
  \city{Shanghai}
  \country{China}}
\email{qzhang@cee.ecnu.edu.cn}

\author{Dan Huang}
\affiliation{
  \institution{Fudan University Shanghai Cancer Center}
  \city{Shanghai}
  \country{China}}
\email{danhuang@shca.org.cn}

\author{Boxiang Yun}
\affiliation{
  \institution{East China Normal University}
  \city{Shanghai}
  \country{China}}
\email{boxiangyun@stu.ecnu.edu.cn}

\author{Hongkai Xiong}
\affiliation{
  \institution{East China Normal University}
  \city{Shanghai}
  \country{China}}
\email{hkxiong@cee.ecnu.edu.cn}

\author{Qingli Li}
\authornotemark[1]
\affiliation{
  \institution{East China Normal University}
  \city{Shanghai}
  \country{China}}
\email{qlli@cs.ecnu.edu.cn}

\renewcommand{\shortauthors}{Siyuan Xu et al.}

\begin{abstract}
Histopathological examination primarily relies on hematoxylin and eosin (H\&E) and immunohistochemistry (IHC) staining. Although IHC provides critical molecular information, it is costly and requires specialized expertise. Stain transfer provides an efficient alternative by computationally generating IHC from H\&E images, but remains challenged by unified and interpretable modeling for heterogeneous biomarkers under pixel-unaligned supervision. We propose DMCoStain, a novel \textbf{D}ata-\textbf{M}odel \textbf{Co}-optimization framework for \textbf{Stain} transfer. It iteratively co-refines training data and model capability, improving staining accuracy and interpretability in both pathological and structural consistency. To refine training data in a clinically meaningful manner, it incorporates the Multimodal Expert-Guided Finer Selection (MEGFS) strategy, built upon a pioneering IHC-positive-expression (IPE) vision-language model (VLM) that emulates pathologist reasoning. To support MEGFS, we construct ImmunoInstruction, the first large-scale IPE instruction-following dataset with 150K VQA samples. Extensive experiments on multiple tissues and biomarkers demonstrate that DMCoStain achieves state-of-the-art (SOTA) accuracy. This paradigm offers strong practical value, and MEGFS also functions as a specialized evaluation tool for future model development. Dataset, code, and more details are in https://github.com/SikangSHU/DMCoStain.
\end{abstract}




\maketitle

\section{Introduction}
\begin{figure}[t]
  \centering
\includegraphics[width=0.47\textwidth]{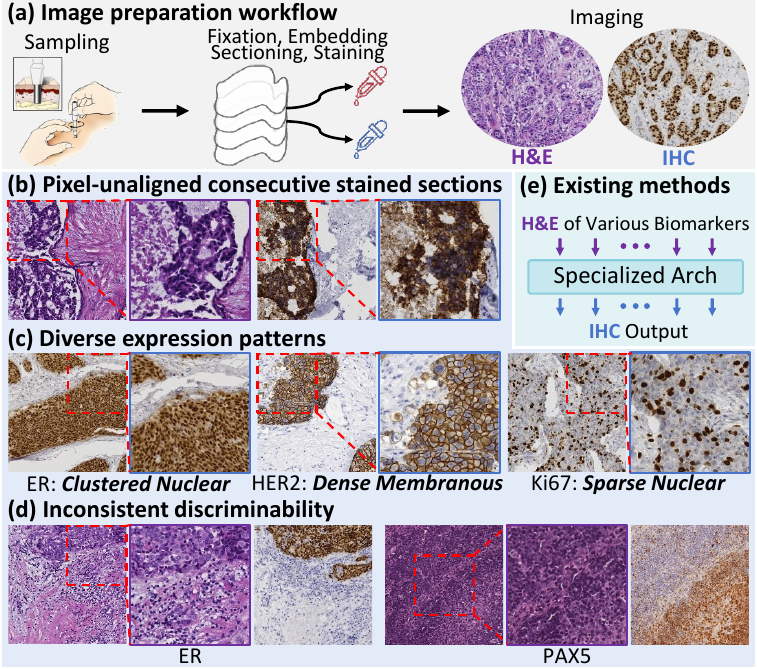}
\vspace{-1.1em}
  \caption{(a) Workflow for H\&E and IHC image preparation; (b-d) Fundamental challenges in H\&E-to-IHC stain transfer; (e) Architecture of existing methods. “Arch” = “Architecture”.}\label{fig_1_1}
  \vspace{-1.8em}
\end{figure}

Cancer remains one of the most serious global health challenges \cite{li2025mico, li2026universal}. Histopathological examination is the gold standard for diagnosis and treatment, providing essential microscopic insights.

In routine pathology, hematoxylin and eosin (H\&E) staining is the most commonly used technique for visualizing tissue architecture and cellular morphology. Hematoxylin stains nuclei blue to dark purple, while eosin colors the cytoplasm and extracellular matrix pink. However, H\&E cannot reveal specific protein expression, which is critical for tumor interpretation. This limitation is addressed by immunohistochemistry (IHC), a molecular-level technique that visualizes protein (positive) expression via antigen-antibody reactions. IHC typically employs diaminobenzidine (DAB) as a chromogen with hematoxylin counterstaining to detect clinically relevant biomarkers (e.g., ER, PR, Ki67, HER2 in breast cancer). Despite its diagnostic value, IHC is time-consuming, costly, and requires specialized expertise, limiting its widespread use. These constraints have motivated stain transfer \cite{klockner2025h}, which computationally generates biomarker-specific IHC images from standard H\&E slides. It is enabled by the intrinsic correlation between tissue morphology and molecular expression, whereby cellular structures and contextual patterns encode underlying biomarker status \cite{liu2021unpaired, shamai2022deep, el2024regression}.

We seek to learn a deep learning-based mapping from H\&E to IHC that enables multi-biomarker diagnosis on the same tissue while preserving pathological and structural information in H\&E. Despite recent progress, several fundamental challenges remain: (1) \textbf{Absence of pixel-aligned ground truth (GT)}. In clinical practice, staining is irreversible, as repeated destaining and restaining of the same section cause chromogenic residue and tissue degradation. Thus, H\&E and IHC staining are performed on consecutive sections, inevitably introducing intrinsic and preparation-induced misalignment (Fig. \ref{fig_1_1}(a, b)). This process achieves only region-level alignment, whose quality varies across datasets, while true pixel-level alignment is unavailable. Most existing methods rely on pixel-unaligned (i.e., weakly paired) data. Some studies \cite{boyd2022region, pati2024accelerating} incorporate region- or cell-level expert annotations for stronger supervision, but such annotations are limited, coarse, and not scalable. (2) \textbf{Challenges in unified modeling under morphological heterogeneity}. Morphological heterogeneity arises in two key aspects. First, biomarkers differ in expression patterns: e.g., HER2 in breast cancer presents dense membranous staining, while Ki67 exhibits sparse nuclear localization. Second, biomarker discriminability in H\&E is inconsistent: ER in breast cancer is relatively apparent, while PAX5 in lymphoma is hard to discern (Fig. \ref{fig_1_1}(c, d)). This heterogeneity makes it difficult for a single model to capture a unified mapping across tissues and biomarkers. Existing models, often biased by task-specific architectures (Fig. \ref{fig_1_1}(e)), exhibit inconsistent performance across datasets. (3) \textbf{Limited practical reliability}. Although existing models can accommodate weakly paired data \cite{guan2025ot}, their underlying feature learning remains opaque. Combined with pixel-unaligned supervision, the lack of interpretability in learning undermines confidence in practical reliability. While such models may improve average metrics, their practical value remains limited.

We present \textbf{\textit{DMCoStain}} (Fig. \ref{fig_3_1}), an \textbf{\textit{iterative data-model co-optimization framework}} that jointly improves training samples and model capability. On the model side, instead of relying on a single architecture for heterogeneous biomarkers, DMCoStain employs biomarker-specific models whose staining capability progressively improves through multi-stage (i.e., iterative) training on refined data (for Challenge~2). On the data side, to overcome the unavailability of pixel-aligned pairs, DMCoStain starts from weakly paired samples and iteratively refines them via inference, coarse-to-fine evaluation, and selection (for Challenge~1). The accumulated better-paired samples guide models to capture precise pathological features, thereby enhancing and stabilizing modeling. This paradigm improves staining accuracy while maintaining interpretability through explicit optimization procedures (for Challenge~3).

\begin{figure}[t]
  \centering
\includegraphics[width=0.47\textwidth]{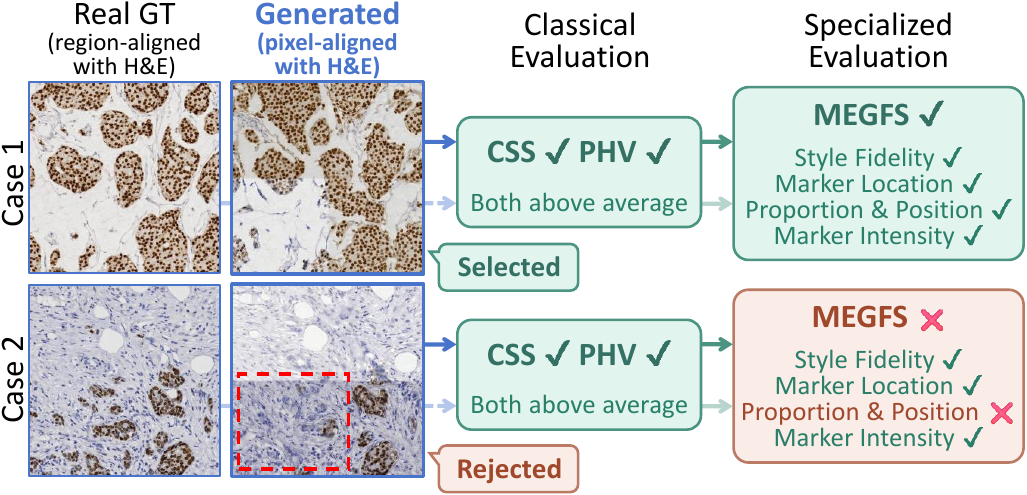}
\vspace{-1.1em}
  \caption{Classical evaluation fails to detect fine-grained local errors. Both cases pass coarse filtering, yet Case 2 contains clinically critical errors in biomarker proportion and position (red box), correctly identified and rejected by MEGFS.}\label{fig_1_2}
  \vspace{-1.8em}
\end{figure}

Specifically, we first train a set of cutting-edge specialized models on private or public weakly paired datasets. These models serve as candidate generators, producing pixel-aligned (i.e., better-paired) virtual IHC images. Inference on the same samples then yields candidate images for selection. Classical evaluation metrics, including Contrast-Structure Similarity (CSS) and Perceptual Hash Value (PHV), measure only image-level similarity between generated images and GT. While capable of identifying obviously low-quality results, they remain insensitive to fine-grained local regions. A generated image with only minor but clinically critical errors in biomarker spatial position or intensity may still meet the fixed thresholds, offering no guarantee of pathological consistency and thus being suitable only for coarse filtering. To enable fine-grained, pathology-aware selection with region-level aligned GT, we propose \textbf{\textit{Multimodal Expert-Guided Finer Selection (MEGFS)}}, centered on the \textbf{\textit{VL Expert-Guided Assessment model (VLEGA)}}, a vision-language model (VLM) inspired by LLaVA \cite{llava, seyfioglu2024quilt} for IHC-positive-expression (IPE) assessment. VLEGA identifies biomarker expression and answers clinically relevant questions for both generated (fake) images and (real) GT. MEGFS evaluates images' consistency across four pathologist-determined aspects: style fidelity, (bio)marker location, marker proportion and spatial position, and marker intensity. Image quality is determined by response consistency, enabling MEGFS to capture local pathological details aligned with clinical practice (Fig. \ref{fig_1_2}). The selected better-paired samples then replace the original weakly paired data, guiding models towards more precise features of biomarkers. DMCoStain iterates this process over multiple stages. To support VLEGA, we construct \textbf{\textit{ImmunoInstruction}}, a large-scale IPE VQA dataset spanning two tissue types (breast cancer and lymphoma) and six biomarkers. It comprises 29,474 IHC images and 147,370 QA pairs across four evaluation categories, with all images sourced from public datasets \cite{ASP, PPT}. All answers are curated and refined by expert pathologists, which directly motivates the term “expert-guided” in MEGFS.

The main contributions of this paper are as follows:
\begin{adjustwidth}{-0.27cm}{}
\begin{itemize}
\item We propose \textit{DMCoStain} to improve staining accuracy in an interpretable manner. It is an \textit{iterative and explicit data-model co-optimization framework} that jointly refines weakly paired samples and biomarker-specific modeling, addressing fundamental challenges in stain transfer and serving as a unifying tool for diverse models likely to emerge in the future.
\item Based on region-level aligned GT, \textit{MEGFS} is proposed to enhance image-level evaluation by assessing staining quality via finer, clinically grounded criteria. Central to it is \textit{VLEGA}, the first IHC-positive-expression VLM that emulates pathologist reasoning.
\item We build \textit{ImmunoInstruction}, the first large-scale IHC-positive-expression VQA dataset, to support VLEGA training. It comprises 30K IHC images and 150K QA pairs spanning diverse biomarkers.
\item Extensive experiments on six biomarkers from breast cancer and lymphoma demonstrate DMCoStain’s SOTA performance in pathology, structure, and style, while varied protocols confirm the framework’s effectiveness across diverse models.

\end{itemize}
\end{adjustwidth}

\section{Related Works}

\subsection{H\&E-to-IHC Stain Transfer}

Early studies focus on color mapping, such as channel-level distribution matching \cite{reinhard2002color} and stain vector decomposition \cite{macenko2009method}, which are limited to color transfer and cannot model pathological relationships. Recent deep learning-based generative methods better capture semantic relationships between stained pairs. Pixel-supervised methods typically adopt Pix2pix \cite{isola2017image} as the backbone, for example the multi-scale pyramid model \cite{pyramidp2p}, but pixel-level losses under imperfect alignment distort tissue structure. To address misalignment, weakly paired training based on CycleGAN \cite{zhu2017unpaired} or CUT \cite{park2020contrastive} is employed. ASP \cite{ASP} introduces an adaptive supervised PatchNCE loss to exploit weak supervision, while PSPStain \cite{pspstain} enhances pathological semantic mining under spatial misalignment. PPT \cite{PPT} designs FocalNCE and patch alignment losses. Diffusion-based transfer is explored in \cite{PST-Diff}, though preserving structure remains challenging. ATST-Net \cite{atstnet} introduces auxiliary task supervision with human-annotation-free masks to ensure pathological consistency and interpretability. More recent methods incorporate optimal transport: SIMGAN \cite{simgan} leverages optimal transport-based supervision with pathological correlation constraints, and USIGAN \cite{peng2026usigan} mitigates weak-pairing effect via unbalanced self-information feature transport and correlation-based consistency mining.

\subsection{MLLMs in Computational Pathology}

Multimodal Large Language Models (MLLMs) have recently advanced computational pathology by enabling joint reasoning over images and text \cite{lu2024visual, ding2025multimodal, xiang2025vision}. PathChat \cite{pathchat} proposes a VL assistant trained on over 456K VL instructions for strong diagnostic QA. PathAsst \cite{pathasst} integrates a pathology-specific CLIP with Vicuna-13B \cite{vicuna2023} and instruction tuning to build a generative foundation model for pathology analysis. CPath-Omni \cite{Cpath-omni}, a 15B-parameter MLLM, unifies patch-level and whole slide image (WSI)-level tasks within a single framework, achieving SOTA performance across diverse benchmarks. SmartPath-R1 \cite{SmartPath-R1} improves reasoning efficiency via reinforcement fine-tuning and a mixture-of-experts model for region of interest (ROI)-level and WSI-level tasks. SlideChat \cite{slidechat} enables direct interaction with WSIs using large-scale instructions, while WSI-LLaVA \cite{Wsi-llava} enhances morphological understanding and explainability through multi-stage training. However, existing MLLMs primarily target complex diagnosis tasks, are large in scale, and lack task-specific, lightweight VLMs tailored to IPE assessment.

\section{Method}

\begin{figure*}[t]
  \centering
\includegraphics[width=0.98\textwidth]{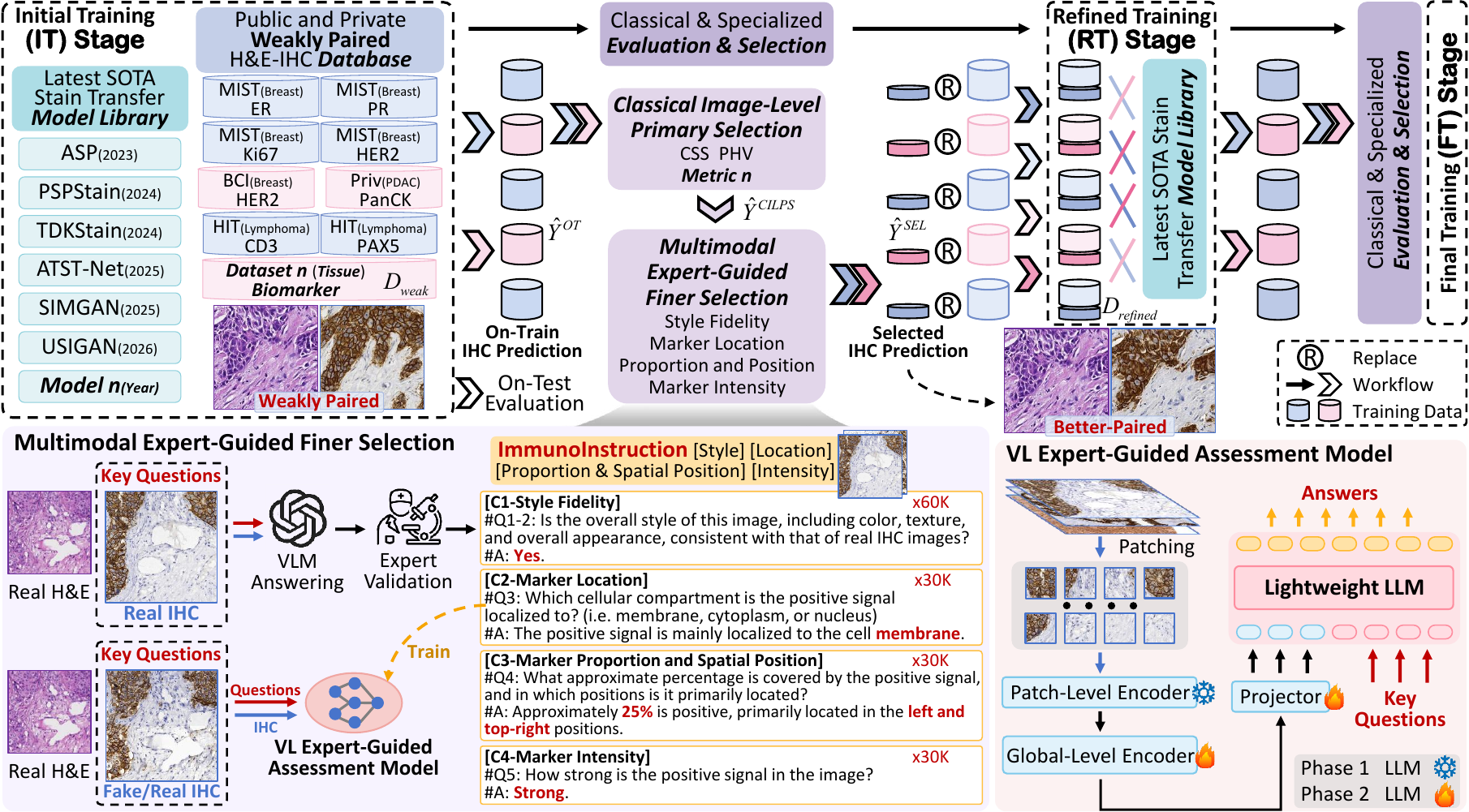}
\vspace{-1em}
  \caption{Overview of the proposed DMCoStain. For training data, darker colors indicate a higher proportion of better-paired samples. “On-Train” and “On-Test” denote inference on training and test data, respectively. Best viewed at a zoomed-in level.}\label{fig_3_1}
  \vspace{-0.8em}
\end{figure*}

\subsection{Overall Architecture of DMCoStain}\label{Overall Architecture}

DMCoStain (Fig. \ref{fig_3_1}) consists of three training stages: \underline{I}nitial \underline{T}raining (IT), \underline{R}efined \underline{T}raining (RT), and \underline{F}inal \underline{T}raining (FT), along with two \underline{e}valuation-and-\underline{s}election (E\&S) stages, forming a progressive pipe\-line: $\text{IT}\xrightarrow{\text{E\&S}}\text{RT}\xrightarrow{\text{E\&S}}\text{FT}$. Each E\&S stage integrates Classical Image-Level Primary Selection (CILPS) and MEGFS.

Given weakly paired H\&E-IHC data $\mathcal{D}_{\text{weak}}= \{(X_i, Y_i)\}_{i=1}^{N}$, a set of cutting-edge stain transfer models $\{M_j\}_{j=1}^{K}$ is first trained in the IT stage, each exhibiting varying accuracy across biomarkers. These models then perform inference on the same training samples to generate On-Train IHC Prediction $\hat{Y}^{\text{OT}}_j = \{M_j(X_i)\}_{i=1}^{N}$, which serve as candidate seeds for data refinement. These predictions are first filtered by CILPS using classical metrics (CSS and PHV) to remove obviously low-quality results, retaining only those satisfying the average-based thresholds of both metrics:
\begin{equation}
    \hat{Y}^{\text{CILPS}} = \left\{ \hat{y} \in \hat{Y}^{\text{OT}} 
    \mid \text{CSS}(\hat{y}) > \bar{\text{CSS}},\ 
    \text{PHV}(\hat{y}) < \bar{\text{PHV}} \right\}.
\end{equation}
Since CILPS is coarse and insensitive to fine local patterns, $\hat{Y}^{\text{CILPS}}$ is further refined by MEGFS, which evaluates clinically relevant aspects through a multimodal, question-driven paradigm, producing Selected IHC Prediction $\hat{Y}^{\text{SEL}} = \text{MEGFS}(\hat{Y}^{\text{CILPS}})$. These are treated as better-paired samples and replace the original weak pairs to form the refined dataset $ \mathcal{D}_{\text{refined}} = \{ (X_i, \hat{Y}^{\text{SEL}}_i \text{ if exists else } Y_i) \mid (X_i, Y_i) \in \mathcal{D}_{\text{weak}} \}$. $\mathcal{D}_{\text{refined}}$ is used in RT to provide higher-quality supervision, guiding models to better capture pathological features from H\&E images. The same E\&S is applied after RT, followed by FT. For efficiency, MEGFS is applied only to the model with the best CILPS performance per biomarker, as CILPS identifies the candidate model most likely to yield a higher proportion of qualified samples, whose predictions are then subjected to fine-grained pathological selection by MEGFS.

While all candidate models participate in IT and RT, only the top-performing model for each biomarker is employed in FT. Since FT requires models to capture precise pathological features while fully preserving H\&E tissue structure, model selection at this stage incorporates an additional structural preservation constraint on top of CILPS. Specifically, the FT model is selected as:
\begin{subequations}
\begin{gather}
    M^{\text{FT}}_b = \arg\max_{M_j \in \mathcal{M}^{\text{valid}}_b} 
        \text{CILPS}_{\text{RT}}(M_j, b), 
    \label{eq:ft_model_selection} \\
    \mathcal{M}^{\text{valid}}_b = \left\{ M_j \,\middle|\, 
        \frac{1}{N_b}\sum_{i=1}^{N_b}
        \text{SSIM}\!\left(\text{gray}(M_j(X_i)),\, X_i^{\text{gray}}\right) 
        \geq \tau_{\text{struct}} \right\},
    \label{eq:valid_model_set}
\end{gather}
\end{subequations}
where $M^{\text{FT}}_b$ is the FT model for biomarker $b$, 
and $\text{CILPS}_{\text{RT}}(M_j, b)$ is the CILPS score of $M_j$ on biomarker $b$ after RT. $\mathcal{M}^{\text{valid}}_b$ denotes the subset of candidate models satisfying the structural preservation constraint, where $\text{gray}(\cdot)$ is grayscale conversion and $X_i^{\text{gray}} = \text{gray}(X_i)$ is the grayscale H\&E image. $\tau_{\text{struct}}$ is the structural preservation threshold for excluding structurally distorted models, and $N_b$ is the number of training pairs for biomarker $b$. Overall, IT selects models that better associate weakly paired data, and RT further identifies models better at handling such data and exploiting better-paired samples for feature extraction. Selected models are finally adopted in FT with accumulated better-paired samples.

As data optimization progresses across stages, the growing availability of high-quality H\&E-IHC pairs gradually reduces reliance on weak supervision, leading to consistent model performance gains. On-Test Evaluation in Sec. \ref{Ablation Study} confirms this trend, with models improving progressively across successive stages.

\subsection{Multimodal Expert-Guided Finer Selection}

MEGFS is illustrated in the lower-left part of Fig. \ref{fig_3_1}. During the construction of ImmunoInstruction, we integrate preliminary responses generated by a general-purpose VLM with expert pathological refinement. Based on it, we develop VLEGA to replace the original “VLM Answering + Expert Validation” workflow. Given a fake or real IHC image, VLEGA answers a predefined set $Q$. A generated image is retained as $\hat{Y}^{\text{SEL}}$ only if all its answers are consistent with those of the corresponding real image:
\begin{equation}
    \hat{Y}^{\text{SEL}} = \left\{ \hat{y} \mid \forall q \in Q,\ 
    \text{VLEGA}(\hat{y}, q) = \text{VLEGA}(y, q) \right\},
\end{equation}
where $y$ is the real image. Training on ImmunoInstruction enables VLEGA to learn robust semantic representation of IPE patterns.

VLEGA adopts an improved LLaVA-style \cite{llava} architecture, consisting of four main components: a patch-level encoder, a global-level encoder, a projection layer, and a lightweight LLM. IHC images, along with a portion of noisy data (see Sec. \ref{ImmunoInstruction}), are first divided into smaller 224$\times$224 patches for computational efficiency. Patch-level features are extracted applying PathoDuet \cite{pathoduet}, a SOTA pretrained model for IHC images based on ViT, to capture fine-grained local features. These patch tokens are then aggregated by a vanilla transformer-based global-level encoder to model global contextual relationships. A projection layer maps visual tokens into an LLM-compatible embedding space, where they are fused with textual question embeddings to generate accurate answers. Benefiting from inherent consistency of IPE characteristics, including color pattern and cellular compartment localization across tissues and biomarkers, VLEGA trained on diverse IHC data generalizes effectively without biomarker-specific retraining, as confirmed in Sec. \ref{Modules in VLEGA}.

\begin{table}[t]
\centering
\fontsize{8.5pt}{9pt}\selectfont
\setlength{\tabcolsep}{7pt}
\renewcommand{\arraystretch}{0.6}
\caption{Statistics of the ImmunoInstruction dataset. HBC = human breast cancer, CL = canine lymphoma; Nuc. / Mem. = nuclear / membranous staining. Each image contains five QA pairs: two for Category~1 and one for each of Categories~2-4.}
\label{tab1}
\vspace{-1.4em}
\begin{tabular}{l c c c c}
\toprule
Dataset\textsubscript{Marker} & Tissue & Location & IHC Image & Q-A/img \\
\midrule
MIST\textsubscript{ER}   & HBC & Nuc. & 4,093 & 5 \\
MIST\textsubscript{PR}   & HBC & Nuc. & 4,134 & 5 \\
MIST\textsubscript{Ki67} & HBC & Nuc. & 4,334 & 5 \\
MIST\textsubscript{HER2} & HBC & Mem. & 4,642 & 5 \\
HIT\textsubscript{PAX5}  & CL  & Nuc. & 6,043 & 5 \\
HIT\textsubscript{CD3}   & CL  & Mem. & 6,228 & 5 \\
\midrule[0.06em]
Total & -- & -- & 29,474 & 147,370 \\
\bottomrule
\end{tabular}
\vspace{-1.6em}
\end{table}

\subsection{ImmunoInstruction}\label{ImmunoInstruction}

The scarcity of large-scale IHC pathology datasets limits VLMs from learning IHC expression patterns. To address this issue, we construct ImmunoInstruction, with images collected from the training splits of public MIST \cite{ASP} and HIT \cite{PPT}, covering two tissue types and six (bio)markers. In total, ImmunoInstruction comprises 29,474 1024$\times$1024 IHC images, with detailed statistics in Tab. \ref{tab1}.

We first exclude images with obvious artifacts from staining, sectioning, or imaging. GPT-4o~\cite{gpt4} generates initial answers by following predefined instructions, yielding responses that cover the required key information with relatively consistent formatting while retaining natural linguistic diversity. These answers are then reviewed and refined by four pathology experts to ensure clinical accuracy and relevance. Each image is annotated under four instruction categories (C) with five questions (Q) in total: (C1, Q1-2) \textit{Style Fidelity}: evaluates whether the overall style matches real IHC images. Since real IHC images inherently satisfy, we introduce “noisy” negative samples by randomly selecting H\&E, immunofluorescence (IF), or heavily blurred IHC images, with answers set to “No” to enhance model discriminability. These three negative categories serve complementary purposes: H\&E and IF images, which differ markedly in appearance from IHC, promote robustness against varying color distribution shifts, while heavily blurred IHC images, sharing the same staining modality but with severely degraded quality, are the most challenging negatives by requiring the model to distinguish obvious degradation from intact IHC appearance. (C2, Q3) \textit{Marker Location}: identifies the cellular compartment of positive expression (i.e., membrane, cytoplasm, or nucleus). (C3, Q4) \textit{Marker Proportion and Spatial Position}: assesses the approximate proportion and spatial distribution of positive regions within the image. (C4, Q5) \textit{Marker Intensity}: evaluates the staining strength of positive signal. This design yields 58,948 QA pairs for \textit{Style Fidelity} and 29,474 pairs for each of the \textit{other three categories}. For data splitting, 80\% of images from MIST$_{\text{ER, PR, Ki67}}$ (nuclear) and HIT$_{\text{CD3}}$ (membranous) are jointly used for training and 20\% for testing. Images from MIST$_{\text{HER2}}$ (membranous) and HIT$_{\text{PAX5}}$ (nuclear) are entirely held out as external test data to assess generalization to unseen biomarkers. Complete details are provided in Sec. 1 of the supplementary materials. Stratified sampling is adopted to maintain the 80/20 ratio across all answer classes of all questions within each biomarker. Notably, we preserve the natural data imbalance to reflect true clinical prevalence and avoid bias from artificial rebalancing. Since IPE recognition of style, position, and intensity is not biomarker-specific, scarce minority answer classes in one biomarker can be compensated by instances from others.

\subsection{Training and Testing Protocol of VLEGA}\label{Training and Testing Protocol of VLEGA}

\subsubsection{Training Phase 1: Multimodal Pre-Alignment} VLEGA training comprises two phases. The first phase aligns textual embeddings with visual representations, enabling the LLM to correctly interpret visual tokens and stabilize subsequent instruction learning. Only Category~1 questions from ImmunoInstruction are used, resulting in 30.1K VQA samples. This phase is formulated as a binary classification task that distinguishes real IHC images from noisy data. Only the global-level encoder and projection layer are updated.

\subsubsection{Training Phase 2: Visual Instruction Learning} In Phase 2, VLEGA learns to integrate and understand features from the two aligned modalities to recognize IPE and generate professional responses. The remaining three instruction categories (45.1K VQA samples) are used for training, and three main components are jointly updated to ensure adaptability. After this phase, VLEGA acquires robust IHC-specific semantic understanding and is integrated into MEGFS as an offline evaluator without further fine-tuning.

\subsubsection{Testing} Predicted answers are evaluated against GT using category-specific rules. For Categories~1, 2, and 4, keyword-level exact matching is adopted. Let $A_{\mathrm{pred}}$ and $A_{\mathrm{gt}}$ denote the extracted keyword from the predicted and GT answers. A prediction is regarded as correct if $\mathbb{I}_{\text{cat}\{1,2,4\}} =1$ when $A_{\mathrm{pred}} = A_{\mathrm{gt}}$, and $0$ otherwise. For Category~3, the evaluation is split into two aspects. For Proportion, since exact matching would be overly strict given the continuous nature of percentage estimates, a tolerance-based consistency rule is applied. Proportion levels are discretized into predefined intervals indexed by $k$, with $\tau_k$ denoting the tolerance threshold for the $k$-th interval. Let $p_{\mathrm{pred}}$ and $p_{\mathrm{gt}}$ denote the predicted and GT proportion levels. A prediction is considered consistent if $\mathbb{I}_{\mathrm{prop}} = 1$ when $\left| p_{\mathrm{pred}} - p_{\mathrm{gt}} \right| < \tau_k$, and $0$ otherwise. For Position, semantic consistency between predicted and GT descriptions is evaluated using a GPT-based similarity \cite{slidechat} function $S_{\mathrm{GPT}}(\cdot,\cdot)$, with threshold $\theta$. Similarity scores range from 1 to 10 in increments of 1, with higher values indicating greater similarity. A spatial description is deemed correct if $\mathbb{I}_{\mathrm{spatial}} = 1$ when $S_{\mathrm{GPT}}\!\left(A_{\mathrm{pred}}, A_{\mathrm{gt}}\right) \ge \theta$, and $0$ otherwise. The advanced GPT-5.1 model is employed for scoring.

In DMCoStain, model training occurs only during the IT, RT, and FT stages, while all E\&S stages operate offline. During E\&S, MEGFS employs VLEGA for evaluation under the same rules. Additional details are provided in Sec. 2 of the supplementary materials.

\label{Comparison with Competitive Methods}

\begin{table*}[t]
\centering
\fontsize{8pt}{9pt}\selectfont
\setlength{\tabcolsep}{8pt}
\renewcommand{\arraystretch}{0.6}
\caption{Quantitative comparison on MIST and HIT. KID is scaled by 1000; bold and \underline{underlined} denote the best and second-best.}
\label{tab_2}
\vspace{-1.4em}
\begin{tabular}{c c | c c | c c c c c c | c c}
\toprule
\multirow{2}{*}{Dataset} &
\multirow{2}{*}{\makecell{IHC Image\\Count}} &
\multirow{2}{*}{\makecell{Method\\(Model)}} &
\multirow{2}{*}{Source} &
\multirow{2}{*}{CSS$\uparrow$} &
\multicolumn{5}{c|}{PHV ($T{=}0.01$)$\downarrow$} & \multirow{2}{*}{FID$\downarrow$} & \multirow{2}{*}{KID$\downarrow$} \\
\cmidrule(lr){6-10}
& & & & & layer1 & layer2 & layer3 & layer4 & avg. & & \\
\midrule

\multirow{10}{*}{\makecell{MIST$_{\text{ER}}$\\Nuc.}} &
\multirow{10}{*}{1000} &
PyramidP2P & CVPRW 22 & 0.097 & 0.477 & 0.465 & 0.363 & 0.852 & 0.539 & 112.7 & 81.5 \\
& & ASP & MICCAI 23 & 0.115 & 0.467 & 0.435 & 0.291 & 0.831 & 0.506 & 68.0 & 23.1 \\
& & PSPStain & MICCAI 24 & \underline{0.139} & 0.494 & 0.435 & 0.286 & 0.825 & 0.510 & \underline{35.5} & 6.7 \\
& & TDKStain & MICCAI 24 & 0.091 & 0.400 & 0.389 & 0.317 & 0.839 & \underline{0.486} & 84.5 & 38.7 \\
& & ATST-Net & IJCAI 25 & 0.112 & 0.495 & 0.440 & 0.288 & 0.826 & 0.512 & 53.4 & 13.8 \\
& & SIMGAN & TMI 25 & 0.121 & 0.473 & 0.435 & 0.285 & 0.834 & 0.507 & 36.6 & 4.8 \\
& & USIGAN & TIP 26 & 0.129 & 0.467 & 0.417 & 0.277 & 0.832 & 0.498 & 35.9 & \underline{4.5} \\
\cmidrule(lr){3-12}
& & DMCoStain & Ours & \textbf{0.155} & 0.414 & 0.365 & 0.241 & 0.801 & \textbf{0.455} & \textbf{32.1} & \textbf{2.8} \\

\midrule

\multirow{10}{*}{\makecell{MIST$_{\text{HER2}}$\\Mem.}} &
\multirow{10}{*}{1000} &
PyramidP2P & CVPRW 22 & 0.080 & 0.465 & 0.441 & 0.333 & 0.841 & 0.520 & 113.0 & 76.6 \\
& & ASP & MICCAI 23 & 0.100 & 0.459 & 0.420 & 0.264 & 0.822 & 0.491 & 54.7 & 14.8 \\
& & PSPStain & MICCAI 24 & 0.106 & 0.515 & 0.454 & 0.281 & 0.825 & 0.519 & 42.7 & \underline{6.5} \\
& & TDKStain & MICCAI 24 & 0.088 & 0.408 & 0.378 & 0.281 & 0.830 & 0.474 & 67.0 & 28.0 \\
& & ATST-Net & IJCAI 25 & 0.095 & 0.482 & 0.430 & 0.270 & 0.818 & 0.500 & 57.3 & 9.5 \\
& & SIMGAN & TMI 25 & 0.090 & 0.480 & 0.434 & 0.277 & 0.825 & 0.504 & 49.0 & 9.7 \\
& & USIGAN & TIP 26 & \underline{0.114} & 0.436 & 0.385 & 0.248 & 0.816 & \underline{0.471} & \underline{39.7} & \textbf{2.3} \\
\cmidrule(lr){3-12}
& & DMCoStain & Ours & \textbf{0.131} & 0.429 & 0.374 & 0.231 & 0.797 & \textbf{0.458} & \textbf{35.2} & \textbf{2.3} \\

\midrule

\multirow{10}{*}{\makecell{HIT$_{\text{PAX5}}$\\Nuc.}} &
\multirow{10}{*}{652} &
PyramidP2P & CVPRW 22 & \underline{0.417} & 0.350 & 0.315 & 0.211 & 0.783 & 0.415 & 64.4 & 19.9 \\
& & ASP & MICCAI 23 & 0.398 & 0.482 & 0.439 & 0.270 & 0.812 & 0.501 & 107.3 & 27.2 \\
& & PSPStain & MICCAI 24 & 0.353 & 0.379 & 0.303 & 0.183 & 0.775 & 0.410 & 51.3 & 3.3 \\
& & TDKStain & MICCAI 24 & \textbf{0.427} & 0.252 & 0.227 & 0.156 & 0.757 & \textbf{0.348} & 54.0 & 5.9 \\
& & ATST-Net & IJCAI 25 & 0.336 & 0.388 & 0.349 & 0.205 & 0.778 & 0.430 & 56.2 & 6.2 \\
& & SIMGAN & TMI 25 & 0.326 & 0.398 & 0.330 & 0.198 & 0.786 & 0.428 & 50.4 & 2.7 \\
& & USIGAN & TIP 26 & 0.336 & 0.351 & 0.288 & 0.186 & 0.779 & 0.401 & \underline{48.6} & \underline{2.2} \\
\cmidrule(lr){3-12}
& & DMCoStain & Ours & 0.352 & 0.328 & 0.262 & 0.158 & 0.751 & \underline{0.375} & \textbf{42.3} & \textbf{1.1} \\

\midrule

\multirow{10}{*}{\makecell{HIT$_{\text{CD3}}$\\Mem.}} &
\multirow{10}{*}{620} &
PyramidP2P & CVPRW 22 & \underline{0.459} & 0.287 & 0.253 & 0.175 & 0.772 & 0.372 & 64.7 & 13.9 \\
& & ASP & MICCAI 23 & 0.435 & 0.399 & 0.338 & 0.207 & 0.787 & 0.433 & 62.5 & 6.3 \\
& & PSPStain & MICCAI 24 & 0.381 & 0.393 & 0.294 & 0.196 & 0.791 & 0.419 & 59.8 & 3.8 \\
& & TDKStain & MICCAI 24 & \textbf{0.476} & 0.221 & 0.183 & 0.132 & 0.754 & \textbf{0.323} & 54.3 & 5.1 \\
& & ATST-Net & IJCAI 25 & 0.334 & 0.454 & 0.371 & 0.198 & 0.733 & 0.439 & 65.0 & 8.9 \\
& & SIMGAN & TMI 25 & 0.406 & 0.369 & 0.308 & 0.197 & 0.779 & 0.413 & 52.6 & \underline{2.3} \\
& & USIGAN & TIP 26 & 0.387 & 0.318 & 0.247 & 0.162 & 0.768 & 0.374 & \underline{48.3} & 2.5 \\
\cmidrule(lr){3-12}
& & DMCoStain & Ours & 0.400 & 0.310 & 0.231 & 0.143 & 0.748 & \underline{0.358} & \textbf{45.2} & \textbf{1.3} \\

\bottomrule
\end{tabular}
\vspace{-1em}
\end{table*}

\begin{table}[t]
\centering
\fontsize{8pt}{9pt}\selectfont
\setlength{\tabcolsep}{7pt}
\renewcommand{\arraystretch}{0.6}
\caption{Segmentation performance of three models across stages on private PDAC$_{\text{CK}}$. “\#Images” = “IHC Image Count”.}
\label{tab_3}
\vspace{-1.4em}
\begin{tabular}{c c c c c c}
\toprule
Dataset & \#Images & Stage & Model & DICE $\uparrow$ & IOU $\uparrow$ \\ \midrule
\multirow{11}{*}{\makecell{PDAC$_{\text{CK}}$\\Mem.}} & 
\multirow{11}{*}{60} & 
\multirow{3}{*}{IT} & ATST-Net & 0.592 & 0.450 \\ 
 &  &  & SIMGAN  & 0.595 & 0.476 \\ 
 &  &  & USIGAN  & 0.648 & 0.532 \\ \cmidrule{3-6}
 &  & \multirow{3}{*}{RT} & ATST-Net & 0.637 & 0.513 \\
 &  &  & SIMGAN  & 0.651 & 0.530 \\
 &  &  & USIGAN  & 0.663 & 0.544 \\ \cmidrule{3-6}
 &  & \multirow{3}{*}{FT} & ATST-Net & \textbf{0.684} & \textbf{0.561} \\
 &  &  & SIMGAN  & 0.673 & 0.550 \\
 &  &  & USIGAN  & 0.670 & 0.544 \\ 
\bottomrule
\end{tabular}
\vspace{-1.6em}
\end{table}

\section{Experiments and Analysis}\label{Experiments and Analysis}

\subsection{Experimental Setup}
For \textbf{\textit{datasets}}, we apply two high-quality public datasets of consecutive H\&E and IHC sections: MIST \cite{ASP} and HIT \cite{PPT}. MIST contains four biomarkers (ER, PR, Ki67, HER2), each with 1,000 test images, while HIT includes PAX5 and CD3 with 652 and 620 test images, respectively. All images have a resolution of 1024$\times$1024. For the segmentation task evaluation in the ablation study, we additionally use a private Pancreatic Ductal Adenocarcinoma (PDAC) dataset of consecutive H\&E and cytokeratin (CK)-IHC sections, comprising 89 training and 15 testing pairs at 40$\times$ magnification and 1024$\times$1024 resolution. Expert pixel-wise annotations are provided on H\&E images, guided by the corresponding CK-IHC sections to delineate CK-positive regions. For \textbf{\textit{evaluation metrics}}, we adopt four metrics \cite{atstnet}: two image-level metrics (CSS, PHV) and two set-level metrics, Fréchet Inception Distance (FID) and Kernel Inception Distance (KID). FID and KID measure feature-space distribution similarity between image sets. For \textbf{\textit{implementation details}}, the model library of DMCoStain comprises PyramidP2P \cite{pyramidp2p}, ASP \cite{ASP}, PSPStain \cite{pspstain}, TDKStain \cite{tdkstain}, ATST-Net \cite{atstnet}, SIMGAN \cite{simgan}, and USIGAN \cite{peng2026usigan}, all of which also serve as baselines for comparison. Whether used independently or within DMCoStain, all models follow their original training settings with a unified epoch schedule. All stain transfer models are trained and inferenced at 512$\times$512 resolution without image normalization, and the resulting patches are subsequently stitched back to 1024$\times$1024 resolution for evaluation and selection. The structural preservation threshold $\tau_{\text{struct}}$ is set to 0.6. Unless otherwise specified, VLEGA employs the relatively lightweight Qwen2.5-3B-Instruct \cite{qwen2} as the LLM. VLEGA is trained in two phases on two 48GB NVIDIA GeForce RTX 4090 GPUs. Phase~1 is trained for 3 epochs with a learning rate (lr) of 0.001, while Phase~2 is trained for 3 epochs with a lr of 0.00002. AdamW is used in both phases. VLEGA is trained and evaluated on 1024$\times$1024 images.

\begin{table}[t]
\centering
\fontsize{8pt}{9pt}\selectfont
\setlength{\tabcolsep}{9.5pt}
\renewcommand{\arraystretch}{0.6}
\caption{Pathologist subjective evaluation scores for USIGAN-generated images across stages on MIST and HIT test data.}
\label{tab_4}
\vspace{-1.4em}
\begin{tabular}{c|cccc}
\toprule
Stage & MIST$_{\text{ER}}$ & MIST$_{\text{HER2}}$ & HIT$_{\text{PAX5}}$ & HIT$_{\text{CD3}}$ \\
\midrule
IT & 6.68 & 6.77 & 6.96 & 7.64 \\
RT & 7.06 & 7.35 & 7.43 & 8.15 \\
FT & \textbf{7.65} & \textbf{7.96} & \textbf{8.06} & \textbf{8.92} \\
\bottomrule
\end{tabular}
\vspace{-1.6em}
\end{table}

\begin{figure*}[t]
  \centering
\includegraphics[width=0.97\textwidth]{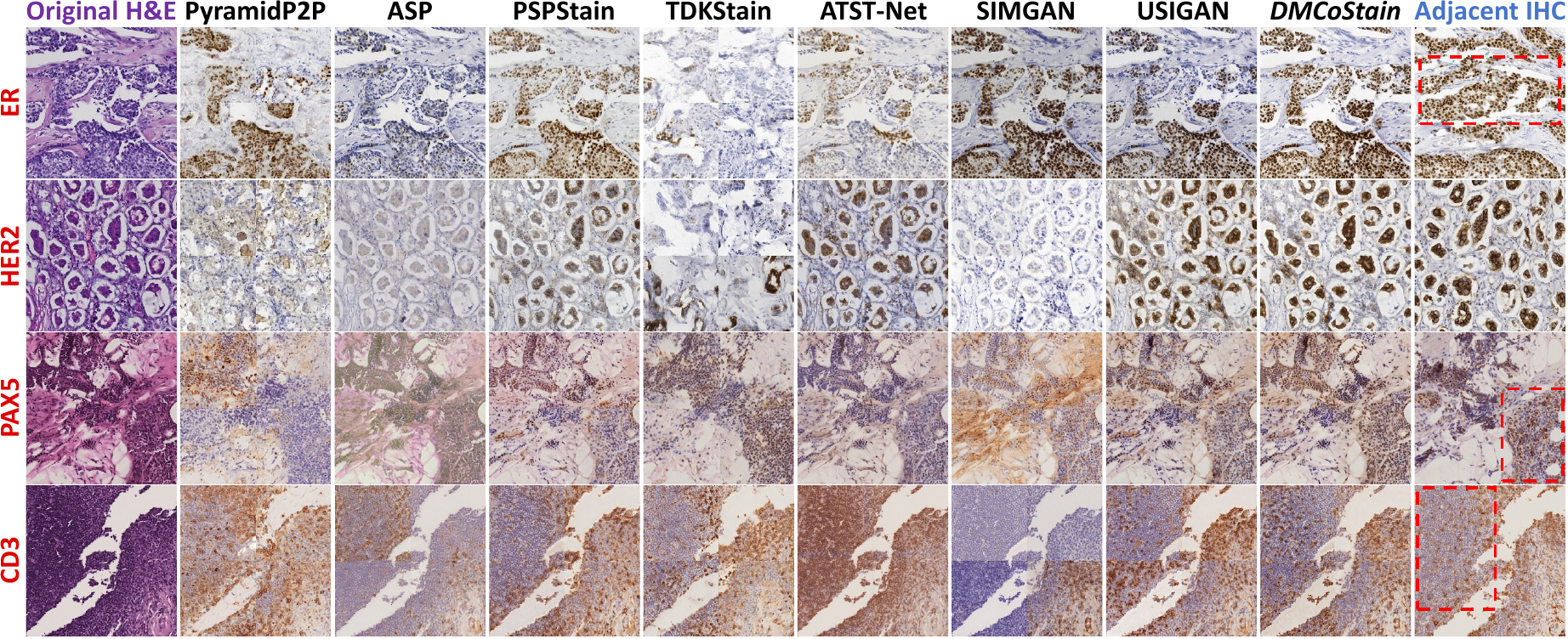}
\vspace{-1.2em}
  \caption{Qualitative comparison on MIST and HIT. Rows show ER, HER2, PAX5, and CD3 from top to bottom, at 1024$\times$1024.}\label{fig_4_1}
  \vspace{-0.8em}
\end{figure*}

\begin{figure}[t]
  \centering
\includegraphics[width=0.42\textwidth]{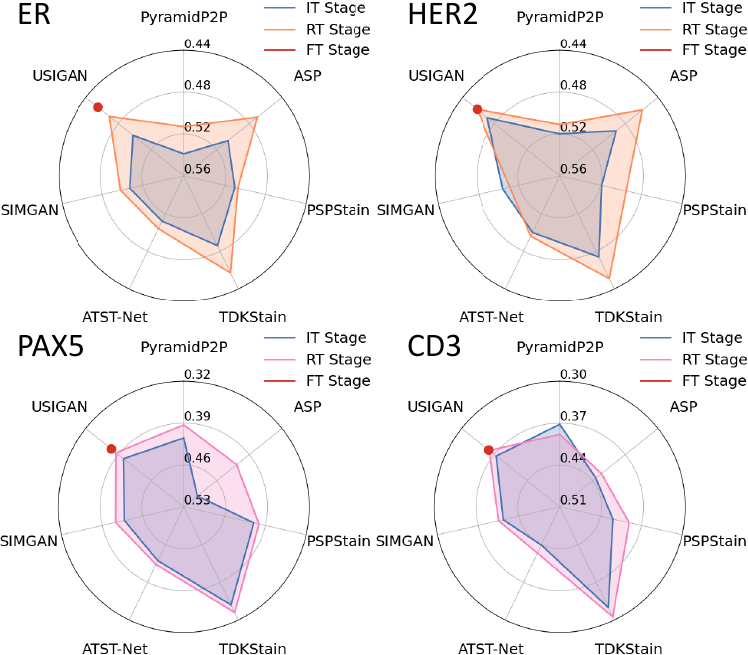}
\vspace{-1em}
  \caption{PHV trends of On-Test Prediction across stages.}\label{fig_4_2}
  \vspace{-1.4em}
\end{figure}

\subsection{Comparison with SOTA Methods}
\subsubsection{Quantitative Comparison} Quantitative results of various methods on two nuclear biomarkers (ER, PAX5) and two membranous biomarkers (HER2, CD3) are presented in Tab. \ref{tab_2}, with more detailed results in Sec. 3 of the supplementary materials. PyramidP2P aligns fake and real IHC feature maps in high-dimensional space using strict constraints on aggregated features. While it preserves partial consistency of positive regions on HIT, where alignment is relatively reliable, it performs poorly on MIST and severely disrupts tissue structure. ASP mitigates the impact of noisy supervision through specialized loss designs, achieving strong performance on MIST. However, its limited representation capacity hinders accurate pathological modeling on HIT, where H\&E morphology is highly similar and biomarker discrimination is more challenging. ATST-Net constructs auxiliary tasks based on masks derived from consecutive IHC sections, but heavy noise in these masks degrades its performance. Notably, TDKStain and USIGAN, benefiting from well-designed task-specific networks, achieve leading CSS and PHV scores across all biomarkers, indicating strong pathological consistency in H\&E-to-IHC style transfer. However, TDKStain's strong loss constraints damage tissue structure, limiting its practical value. In contrast, DMCoStain integrates the complementary strengths of multiple models across biomarkers. By selecting fully aligned generated IHC images to guide subsequent training, it focuses on precise pathological features, markedly improving pathological consistency while fully preserving structure. Consistently superior PHV scores indicate accurate multi-level feature matching between fake and real IHC images, while FID and KID further confirm feature distribution consistency. On MIST$_{\text{ER}}$, DMCoStain improves CSS, PHV (avg.), FID, and KID by 0.016, 0.031, 3.4, and 1.7, respectively.

\begin{figure}[t]
  \centering
\includegraphics[width=0.45\textwidth]{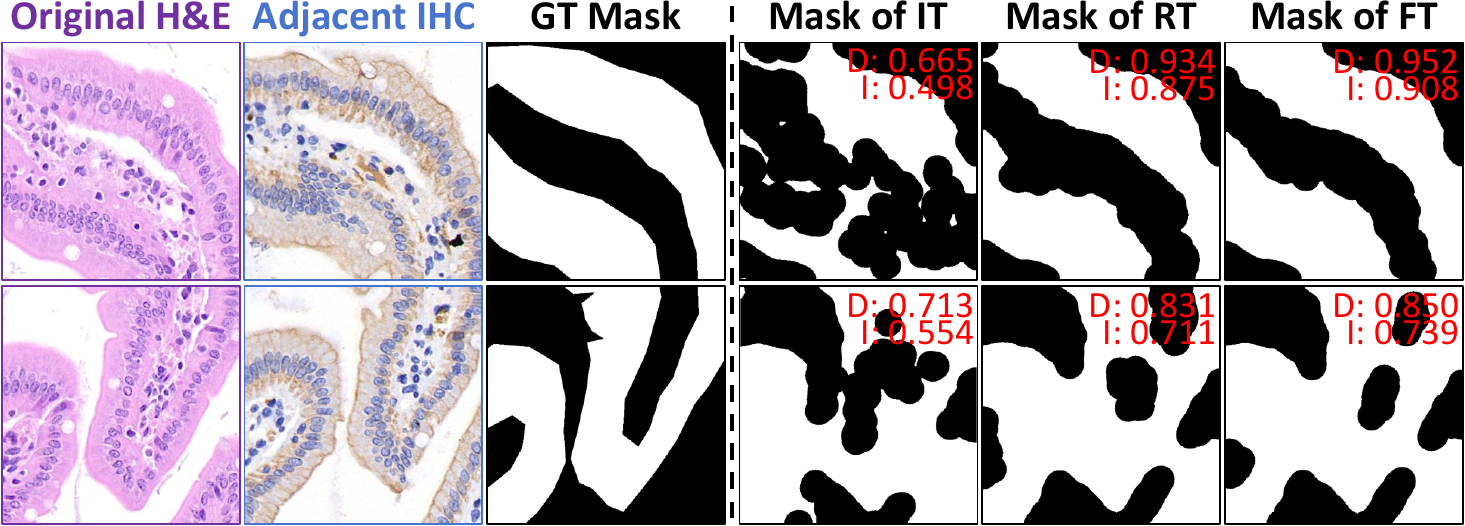}
\vspace{-1.1em}
  \caption{Segmentation results on private PDAC$_{\text{CK}}$ test data across stages. D = Dice, I = IoU. All images are 512$\times$512.}\label{fig_4_3}
  \vspace{-1.4em}
\end{figure}

\subsubsection{Qualitative Comparison} Fig. \ref{fig_4_1} shows qualitative results. PyramidP2P and TDKStain distort the original H\&E structure, while other methods preserve structural integrity. Using consecutive (adjacent) IHC sections as reference, DMCoStain exhibits superior pathological consistency. In the red-boxed regions, both the position and intensity of positive expression in the generated images align more closely with the real IHC GT. DMCoStain also better reproduces IHC staining style. Moreover, patches generated by DMCoStain show minimal discontinuity, indicating accurate capture of intrinsic biomarker-specific features without being affected by local context variations, as is particularly evident in the last row.

\subsection{Ablation Study}\label{Ablation Study}

We analyze On-Train and On-Test Prediction accuracy across DMCoStain's stages to validate the data-model co-optimization framework, further assess it via downstream segmentation and expert evaluation, and finally examine VLEGA’s core design choices.

\subsubsection{Stage-Wise Performance of DMCoStain} After the RT stage, USIGAN is identified as the best-performing model across all four biomarkers and is adopted as the sole training model in FT. Although TDKStain achieves the highest raw CILPS scores on HIT$_{\text{PAX5}}$ and HIT$_{\text{CD3}}$, it is automatically excluded by the structural preservation constraint due to severe tissue distortion, consistent with the model selection process defined in Sec. \ref{Overall Architecture}. The numbers of Selected IHC Prediction for each biomarker and stage are provided in Sec. 3 of the supplementary materials. Fig. \ref{fig_4_2} shows the PHV trends of On-Test Prediction across stages, with red points indicating the FT stage results of USIGAN (i.e., DMCoStain). Detailed On-Train and On-Test Prediction results are in Sec. 3 of the supplementary materials. Overall, nearly all models exhibit consistent improvement as stages advance. Notably, on HIT$_{\text{PAX5}}$, ASP benefits substantially from better-aligned pairs, as the reduction in noisy samples alleviates reliance on low-quality supervision, yielding an 8.4\% PHV improvement. These results fully demonstrate that the data-model co-optimization framework generates high-quality data and leverages it to enhance model capability in stain transfer.

\subsubsection{Segmentation Task Evaluation} Biomarker-specific stain transfer enables molecular-related region segmentation on H\&E-stained pathological images by providing accurate IHC images as an intermediate representation. Segmentation is performed via stain transfer followed by conventional morphological operations, including color deconvolution \cite{colordeconv}, grayscale conversion, Otsu thresholding, and closing. Since segmentation accuracy correlates positively with staining quality, segmentation improvement across stages serves as an indirect but objective measure of staining quality gains. Here, we evaluate three recent models on the private PDAC$_{\text{CK}}$ dataset using Dice and IoU as metrics. Tab. \ref{tab_3} reports segmentation results on the test data across IT, RT, and FT stages, showing substantial improvement for all models. ATST-Net achieves the largest gains, with Dice increasing by 9.2\% and IoU by 11.1\% from IT to FT, highlighting its particular reliance on better-paired training data. Fig. \ref{fig_4_3} visualizes segmentation masks using ATST-Net as the backbone, showing that predicted masks progressively approach the GT masks from IT to FT. These results show data-model co-optimization's effectiveness.

\begin{table}[t]
\centering
\fontsize{8pt}{9pt}\selectfont
\setlength{\tabcolsep}{3.6pt}
\renewcommand{\arraystretch}{0.6}
\caption{Comparison of LLM backbones and model scales in VLEGA on internal and external test data. Accuracy is 0-100\% for C1, C2, C3(1), and C4; similarity score is 1-10 for C3(2).}
\label{tab_5}
\vspace{-1.4em}
\begin{tabular}{c cc ccccc}
\toprule
LLM & Training & C1 & C2 & C3(1) & C3(2) & C4 \\
\midrule
\multicolumn{7}{c}{\textit{Internal Test Data (above)} and \textit{External Test Data (below)}} \\
\midrule
Random                & -- & 50.3 & 34.0 & -- & -- & 24.9 \\
GPT-4o                & --& 85.8 & 75.6 & 19.9 & 4.43 & 57.4 \\
Qwen2.5-1.5B-Instruct & $\checkmark$ & 79.5 & 73.4 & 63.1 & 6.37 & 72.0 \\
LLaMA-3.2-3B-Instruct & $\checkmark$ & 88.6 & 79.4 & 73.3 & 7.71 & 78.9 \\
\cellcolor[rgb]{0.88,0.88,0.88}{Qwen2.5-3B-Instruct}   & \cellcolor[rgb]{0.88,0.88,0.88}$\checkmark$ & \cellcolor[rgb]{0.88,0.88,0.88}91.0 & \cellcolor[rgb]{0.88,0.88,0.88}81.0 & \cellcolor[rgb]{0.88,0.88,0.88}82.1 & \cellcolor[rgb]{0.88,0.88,0.88}8.62 & \cellcolor[rgb]{0.88,0.88,0.88}80.8 \\
\midrule
GPT-4o                & -- & 84.2 & 70.6 & 25.0 & 3.49 & 56.8 \\
Qwen2.5-1.5B-Instruct & $\checkmark$ & 76.7 & 69.6 & 60.3 & 6.49 & 72.8 \\
LLaMA-3.2-3B-Instruct & $\checkmark$ & 86.7 & 78.9 & 71.9 & 7.18 & 77.0 \\
\cellcolor[rgb]{0.88,0.88,0.88}{Qwen2.5-3B-Instruct}   & \cellcolor[rgb]{0.88,0.88,0.88}$\checkmark$ & \cellcolor[rgb]{0.88,0.88,0.88}89.8 & \cellcolor[rgb]{0.88,0.88,0.88}82.4 & \cellcolor[rgb]{0.88,0.88,0.88}80.7 & \cellcolor[rgb]{0.88,0.88,0.88}7.93 & \cellcolor[rgb]{0.88,0.88,0.88}78.6 \\
\toprule
\end{tabular}
\vspace{-2em}
\end{table}

\subsubsection{Subjective Evaluation} Three pathologists with over five years of clinical experience conduct a blinded evaluation of test results generated by USIGAN (the FT model) at different training stages. To eliminate potential bias, pathologists are blinded to the stage assignment of each image. Original H\&E and corresponding IHC images serve as reference. Images are scored on a 10-point scale (interval = 1), with higher scores indicating better quality. Evaluation criteria include preservation of IHC staining style and accuracy of positive expression regarding marker location, proportion, spatial distribution, and intensity. Scores are defined as: 8-10, no noticeable errors and high consistency with reference, meeting clinical requirements; 6-7, minor errors in some aspects without affecting overall judgment; 4-5, multiple errors and poor overall quality; 1-3, completely incorrect. Three experts independently evaluate 50 non-overlapping image groups per biomarker, with each group comprising one image from each of the three stages, totaling 150 groups per biomarker. Final scores averaged across all images are shown in Tab. \ref{tab_4}. Scores consistently increase across stages, and at FT, all biomarkers approach or exceed the clinical applicability threshold, with HIT$_{\text{CD3}}$ achieving the highest score of 8.92.

\subsubsection{Modules in VLEGA}\label{Modules in VLEGA} We compare different LLM backbones and model scales for our task. Unlike general pathological image diagnosis, which involves diverse instructions and complex tumor microenvironments and thus typically requires medium-scale LLMs \cite{slidechat, Patho-R1, zhang2026patho}, IPE recognition features fixed instructions and relatively simple visual inputs. Accordingly, we prioritize relatively lightweight LLMs to balance accuracy and computational efficiency. Tab. \ref{tab_5} reports accuracy across question categories, with internal testing results in the upper section and external results on biomarkers unseen during training in the lower section. All results are averaged over the test data. As a baseline, candidate keywords in answers are randomly sampled for C1, C2, and C4, approximating the theoretical probability distribution but yielding poor performance. GPT-4o, used in MEGFS to generate preliminary answers during dataset construction, achieves decent accuracy on internal data for C1 and C2 (85.8\% and 75.6\%), but performs poorly on C3 and C4, highlighting the necessity of expert correction when using general-purpose VLMs. Overall, VLEGA with Qwen2.5-3B-Instruct achieves the best performance across all categories, outperforming LLaMA-3.2-3B-Instruct at a comparable scale and substantially surpassing the lighter Qwen2.5-1.5B-Instruct. It achieves 91.0\% accuracy on C1 and exceeds 80\% on all other categories. Given the inherent subjectivity involved in expert correction, this level of performance is comparable to a “VLM Answering + Expert Validation” paradigm. External evaluation shows only marginal degradation on unseen biomarkers, demonstrating strong generalization without any retraining. We further examine the necessity of the two-phase training strategy by incorporating all instruction categories into Phase~1 training. This leads to non-convergence and redundant outputs containing irrelevant information, indicating that the projection layer alone is insufficient to align visual and textual features across diverse instruction categories and that LLM fine-tuning is essential. We additionally remove the global-level encoder under the standard two-phase training, leaving only the projection layer for multimodal pre-alignment. This prevents convergence even in Phase~1, underscoring the critical role of the global-level encoder in integrating patch tokens. These results confirm the necessity of the two-phase training strategy and all core modules.

To assess the computational cost of VLEGA, we measure end-to-end inference time and GPU memory usage. The entire pipeline, from input image to output answer, completes within 0.5s for all question categories, with C1 and C4 taking approximately 0.1s each. GPU memory consumption is 7.2GB, indicating fast and resource-efficient inference well-suited for clinical deployment.

\section{Conclusion}

We propose a new paradigm for stain transfer based on iterative data-model co-optimization. Extensive experiments demonstrate its reliability in both accuracy and interpretability. The complete framework, with its essential evaluation module MEGFS, supports future model development and clinical deployment. By enabling reliable virtual IHC generation from standard H\&E slides, this work reduces dependence on costly molecular staining and facilitates broader clinical adoption of IHC techniques. Overall, it provides a robust pathway towards reliable AI-assisted histopathology.

\begin{acks}
This work is supported by the National Natural Science Foundation of China (Grant Nos. 42527804, 62475072, 62471182, 82572674, and U25A20647), the Science and Technology Commission of Shanghai Municipality (Grant Nos. 25xtcx00600 and 22DZ2229004), the Fundamental Research Funds for the Central Universities, the Shanghai Rising-Star Program (Grant No. 24QA2702100), and the Joint Laboratory of Hyperspectral Big Data and Artificial Intelligence (No. MIP202605).
\end{acks}

\bibliographystyle{ACM-Reference-Format}
\bibliography{sample-base}

\end{document}